\documentclass[10pt,twocolumn,letterpaper]{article}

\usepackage{cvpr}              

\usepackage[dvipsnames]{xcolor}

\usepackage{array}
\usepackage{booktabs}
\usepackage{caption}
\usepackage{subcaption}
\usepackage{pgffor}
\usepackage{pifont}
\newcommand{\cmark}{\ding{51}}
\newcommand{\xmark}{{\color{lightgray}\ding{55}}}

\newcommand{\mysection}[1]{\vspace{2pt}\noindent\textbf{#1}}

\newcommand\blfootnote[1]{%
  \begingroup
  \renewcommand\thefootnote{}\footnote{#1}%
  \addtocounter{footnote}{-1}%
  \endgroup
}

\definecolor{cvprblue}{rgb}{0.21,0.49,0.74}\usepackage[pagebackref,breaklinks,colorlinks]{hyperref}

\usepackage{multirow}
\usepackage{censor}
\usepackage[accsupp]{axessibility}

\usepackage{listings}
\usepackage{xcolor}

\definecolor{codegreen}{rgb}{0,0.6,0}
\definecolor{codegray}{rgb}{0.5,0.5,0.5}
\definecolor{codepurple}{rgb}{0.58,0,0.82}
\definecolor{backcolour}{rgb}{0.95,0.95,0.92}

\lstdefinestyle{mystyle}{
    backgroundcolor=\color{backcolour},
    commentstyle=\color{codegreen},
    keywordstyle=\color{magenta},
    numberstyle=\tiny\color{codegray},
    stringstyle=\color{codepurple},
    basicstyle=\ttfamily\footnotesize,
    breakatwhitespace=false,
    breaklines=true,
    captionpos=b,
    keepspaces=true,
    numbers=left,
    numbersep=5pt,
    showspaces=false,
    showstringspaces=false,
    showtabs=false,
    tabsize=2,
    frame=single,
    framexleftmargin=15pt
}

\def\paperID{17} 
\def\confName{CVsports}
\def\confYear{2024}

\title{
SoccerNet Game State Reconstruction:\\End-to-End Athlete Tracking and Identification on a Minimap}

\author{
Vladimir Somers$^{1,3,8}$* 
\quad Victor Joos$^1$* 
\quad Anthony Cioppa$^{2,4}$*
\quad Silvio Giancola$^4$*
\and Seyed Abolfazl Ghasemzadeh$^1$
\quad Floriane Magera$^{2,7}$
\quad Baptiste Standaert$^1$
\quad Amir M. Mansourian$^5$
\and Xin Zhou$^6$
\quad Shohreh Kasaei$^5$
\quad Bernard Ghanem$^4$ 
\and Alexandre Alahi$^3$ 
\quad Marc Van Droogenbroeck$^2$
\quad Christophe De Vleeschouwer$^1$
\and  $^1$ {\small UCLouvain}
\quad $^2$ {\small University of Liège}
\quad $^3$ {\small EPFL}
\quad $^4$ {\small KAUST}
\quad $^5$ {\small SUT}
\quad $^6$ {\small Baidu Research}
\quad $^7$ {\small EVS}
\quad $^8$ {\small Sportradar}
}

\begin{document}
\maketitle
\begin{abstract}
    Tracking and identifying athletes on the pitch holds a central role in collecting essential insights from the game, such as estimating the total distance covered by players or understanding team tactics. 
    This tracking and identification process is crucial for reconstructing the \emph{game state}, defined by the athletes' positions and identities on a 2D top-view of the pitch, (\ie a \emph{minimap}).
    However, reconstructing the game state from videos captured by a single camera is challenging.
    It requires understanding the position of the athletes and the viewpoint of the camera to localize and identify players within the field.
    In this work, we formalize the task of Game State Reconstruction and introduce \emph{SoccerNet-GSR}, a novel Game State Reconstruction dataset focusing on football videos. 
    \emph{SoccerNet-GSR} is composed of $200$ video sequences of $30$ seconds, annotated with $9.37$ million line points for pitch localization and camera calibration, as well as over $2.36$ million athlete positions on the pitch with their respective role, team, and jersey number.
    Furthermore, we introduce \emph{GS-HOTA}, a novel metric to evaluate game state reconstruction methods. 
    Finally, we propose and release an end-to-end baseline for game state reconstruction, bootstrapping the research on this task.
    Our experiments show that GSR is a challenging novel task, which opens the field for future research.
    Our dataset and codebase are publicly available at \url{https://github.com/SoccerNet/sn-gamestate}.
    \blfootnote{\textbf{(*)} Equal contributions. 
    Data/code available at \href{https://www.soccer-net.org}{www.soccer-net.org}.
    }
\end{abstract}

\section{Introduction}
\label{sec:intro}

Recently, sports companies and teams have shown a growing interest in collecting athlete-centric data. 
One key focus area lies in tracking and identifying athletes on the sports field throughout the entire game, using available video footage.
These analytics hold immense value for a broad spectrum of sports applications, ranging from 
\textbf{(i)}~supporting team coaching and athlete training,
\textbf{(ii)}~assisting scouters in discovering new talents,
\textbf{(iii)}~offering valuable insights for medical staff, and
\textbf{(iv)}~boosting fan engagement through personalized content creation~\cite{Chen2010Personalized,  Chen2011Automatic, Fernandez2010Browsing}.

However, the manual generation of such data by human annotators is time-consuming and costly.
Sensor-based solutions offer a time-efficient alternative, but require athletes to wear special, sometimes expensive, equipment.
Recently, automatic solutions based on optical tracking systems have gained prominence. These systems necessitate the installation of sophisticated, well-calibrated static multi-camera setups in stadiums. Hence, they come with significant drawbacks in terms of cost and scalability, which restricts their use to elite competitions, exemplified by their deployment at events like the 2022 Qatar World Cup.

\begin{figure}[t!]
\centering
\includegraphics[width=0.99\linewidth]{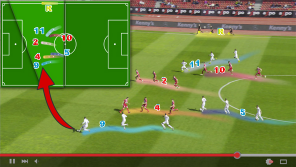}
\caption{
\textbf{SoccerNet-GSR.}
We introduce a novel Game State Reconstruction task, dataset, evaluation metric and baseline.
Our SoccerNet-GSR dataset contains unique identifications for players along with their localization on the pitch, for $200$ video sequences.
  }
  \vspace{-2mm}
\label{fig:pull_figure}
\end{figure}

Meanwhile, recent progress in computer vision opened up a growing potential to automatically and reliably extract athlete localization and identification data solely from broadcast camera feeds.
In line with this objective, Multi-Object Tracking (MOT) methods have long been popular for sports video analysis. 
However, they offer only a partial solution to the aforementioned requirements.
Indeed, the bounding-box-based tracking data produced by MOT (1) lacks critical identification information necessary to analyze specific athletes and (2) lacks interpretability due to the absence of grounding in a real-world coordinate system.
These significant limitations hinder the usability of such tracking data for many downstream sports applications.

To address the above limitations, we introduce the concept of \textbf{G}ame \textbf{S}tate \textbf{R}econstruction (\textbf{GSR}), a novel computer vision task tailored for sports analytics. 
GSR aims to recognize the state of a sports game by identifying and tracking all athletes on the pitch based on input videos captured by a single camera. 
Moreover, game state data can be visualized in a minimap of the game, as depicted in \cref{fig:pull_figure}, offering a concise representation of the ongoing gameplay dynamics.
To support research on this task, we publicly release \emph{SoccerNet-GSR}, the first dataset for Game State Reconstruction, consisting of 200 30-second fully annotated clips.
Our proposed GSR annotations include over 9.37 million line points for football pitch registration, and over 2.36 million athlete positions on the pitch with unique identification information, including their role, team, and jersey number.
Since existing metrics for Multi-Object Tracking~\cite{Bernardin2008Evaluating, Luiten2020HOTA} are not suited for our proposed task, we introduce the \emph{GS-HOTA}, a new evaluation metric to benchmark GSR methods.
Finally, we propose \emph{GSR-Baseline}, the first end-to-end and open-source pipeline for game state reconstruction, built upon state-of-the-art tracking, re-identification, team affiliation, jersey number recognition, pitch localization, and camera calibration methods.
Our analysis underscores the complexity of Game State Reconstruction and highlights the importance of introducing this new benchmark. 
This initiative establishes an ideal platform for future research in the field, aiming to democratize access to this valuable game state data for all leagues.

\mysection{Contributions.} We summarize our contributions as follows. 
\textbf{(i)}~We introduce and concretely define the concept of \emph{Game State Reconstruction}, a task aiming to track and identify all athletes on a minimap of the pitch.
\textbf{(ii)}~We publicly release \textbf{SoccerNet-GSR}, the first open-source sports video dataset for Game State Reconstruction.
\textbf{(iii)}~We introduce \textbf{GS-HOTA}, a new metric to evaluate game state reconstruction methods.
\textbf{(iv)}~We propose \textbf{GSR-Baseline}, the first end-to-end GSR pipeline for football videos.

\section{Related Work}

Game State Reconstruction relates to the general topic of sports video understanding and, more particularly, to tracking, identification, and sports field registration.

\mysection{Sports Video Understanding.}
Sports video understanding has emerged as a prominent research topic over the past decade~\cite{Moeslund2014Computer,Thomas2017Computer,Naik2022AComprehensive, Held2024XVARS}.
Some works focused on low-level semantics, aiming to build a bottom-up understanding of the game~\cite{Cioppa2018ABottomUp}, such as segmenting~\cite{Cioppa2019ARTHuS} or detecting~\cite{Vandeghen2022SemiSupervised,Seweryn2024Improving-arxiv,Rao2015ANovel, Parisot2017SceneSpecific} players and keypoints~\cite{Ludwig2023AllKeypoints,Ghasemzadeh2021DeepSportLab-arxiv}.
Recent advances in computer vision allowed for a higher semantic understanding of the game, focusing for instance on the action spotting task, aiming to spot a series of events during the game~\cite{Cioppa2020AContextaware,Giancola2021Temporally,Zhou2021Feature-arxiv,Zhu2022ATransformerbased,Soares2022Temporally,Cao2022SpotFormer,Giancola2023Towards,Wang2023ABoosted-arxiv,Hong2022Spotting, Leduc2024SoccerNetDepth, Magera2024AUniversal}. 
Fortunately, these works can rely on the availability of large-scale datasets~\cite{Pappalardo2019Apublic,Yu2018Comprehensive,Scott2022SoccerTrack,Jiang2020SoccerDB, Deliege2021SoccerNetv2,Cioppa2022Scaling, Cioppa2022SoccerNetTracking,Mkhallati2023SoccerNetCaption,Held2023VARS} and challenges~\cite{Cioppa2023SoccerNetChallenge-arxiv,Giancola2022SoccerNet,Istasse2023DeepSportradarv2,VanZandycke2022DeepSportradarv1,Midoglu2022MMSys,Kiefer2023macvi}.
Our novel Game State Reconstruction task stands in between low- and high-level semantics, providing both local information about the players but also global information about the whole state of the game through time. 
This information can later be used to better understand player actions~\cite{Cioppa2021Camera, Cabado2024Beyond}, enhance the generation of engaging captions~\cite{Chan2022ToStart,Mkhallati2023SoccerNetCaption}, improve visualizations~\cite{Sacha2017Dynamic,Fischer2019Videobased,ZhuTian2023Sporthesia,Boeker2023Soccer}, or derive high-level analytics~\cite{perin2013soccerstories,decroos2018automatic,Pappalardo2019PlayeRank,Beal2020Optimising-arxiv,Kurach2020Google,ArbuesSanguesa2020Using}. 
In this work, we complement the literature in sports video understanding by proposing a novel task of Game State Reconstruction that aggregates several tasks ranging from field to player understanding.


\mysection{Player Tracking and (Re-)Identification.}
Multiple Object Tracking (MOT) has often been approached through the tracking-by-detection paradigm~\cite{Bewley2016Simple,Veeramani2018DeepSort,Bochinski2018Extending,Bergmann2019Tracking,sun2019deep,Zhang2022Bytetrack,Zhang2021FairMOT,Theiner2021Extraction}. 
However, applying the tracking-by-detection paradigm to sports introduces unique challenges compared to generic scenarios.
Previous works~\cite{Cioppa2020Multimodal,Vandeghen2022SemiSupervised,Cioppa2022SoccerNetTracking,Yang2022TheSecondplace-arxiv,Shitrit2023SoccerNet-arxiv,Huang2023Iterative,Raja2024Tracking} tackled the challenges of similar appearances and fast motion of people and object in sports.
Furthermore, unlike generic MOT scenarios, athletes come in and out of the camera view, requiring long-term Re-Identification (ReID)~\cite{Zhang2022Adaptive,Ghosh2023Relation,Kim2023Feature,Zhang2023PHA, Mansourian2023Multitask}. 
%
Finally, uniquely identifying actors in a sports scene has been widely investigated in the literature. 
Some approaches focused on 
athletes' role (\eg, player, referee, coach, \etc)~\cite{Cioppa2022Scaling, Vandeghen2022SemiSupervised, Mansourian2023Multitask},
their team~\cite{Istasse2019Associative, Mansourian2023Multitask}, 
or jersey numbers~\cite{ye2005jersey,Gerke2015Soccer,Liu2019Pose,Nady2021Player,Vats2021Multitask,Balaji2023Jersey,Liu2023Automated}.
Different from previous works, our new game state reconstruction task combines athlete tracking and identification, including the role, team, and jersey number under a single task.

\mysection{Sports Field Registration.}
Mapping the video tracking data into a real-world coordinate system requires camera calibration. 
Sports games come naturally with a coordinate system based on the sports pitch. 
Hence, combining the location of the field~\cite{Yang2022TheSecondplace-arxiv, SantosMarques2023Hierarchical}  
with video camera calibration~
\cite{Farin2003RobustCamera,Pourreza2008Robust,Sha2020EndtoEnd,Theiner2023TVCalib},
one can reconstruct a game state as illustrated in \cref{fig:pull_figure}.
Unifying tracking and camera calibration as proposed in this paper has been investigated in previous works.\cite{Stein2018Bring,Scott2022SoccerTrack,Karungaru2022Soccer,Cioppa2021Camera}
Scott~\etal~\cite{Scott2022SoccerTrack} collected data from fish-eye camera, drone, and GNSS, while Karungaru~\etal~\cite{Karungaru2022Soccer} focused on the mapping of players onto the field in one video frame. 
Cioppa~\etal~\cite{Cioppa2021Camera} leveraged tracking and camera calibration to reproject players' positions on the pitch for the task of action spotting. 
%
Maglo~\etal~\cite{Maglo2023Individual} introduced a robust player tracking method, incorporating test-time fine-tuning and a novel football field registration technique, which were combined to explore player localization on a minimap.
However, due to the lack of annotations, they did not perform either player identification or quantitative evaluations of their localization results.
Finally, Theiner~\etal~\cite{Theiner2021Extraction} introduced a pipeline to localize players on a pitch minimap from broadcast videos but omitted player identification and tracking. 
Different from previous work, our proposed GSR benchmark addresses the combined athlete pitch localization and identification task.

\section{Game State Reconstruction Task}\label{sec:gsr_task}


Game State Reconstruction (\textbf{GSR}) is a form of video compression task aiming to extract high-level information about the dynamics of a sports game from an input video. 
It includes \textbf{(1)} the 2D position of all athletes on the sports pitch, \textbf{(2)} their roles in the game (e.g., ``player'', ``goalkeeper'', or ``referee''), and \textbf{(3)}, for players, their jersey number and team affiliation.
This information can be visualized on a 2D top-view of the pitch, or minimap, as illustrated in \cref{fig:pull_figure}. 
In the following, we refer to all individuals to be identified and localized, irrespective of their specific roles, as ``\textbf{athletes}''. 
GSR is a multifaceted task that requires addressing various intricate sub-tasks, including: \textbf{(a)} pitch localization and camera calibration, \textbf{(b)} athlete detection, re-identification, and tracking, and \textbf{(c)} role classification, team affiliation, and jersey number recognition.

We formalize the Game State Reconstruction task as follows.
Given a team sports video composed of $T$ frames, the objective is to predict a set of detections \(d_i^t\) for each frame \(t\), where \(i\) indexes the detections within frame \(t\). 
A detection encapsulates each athlete's location on the pitch (\(pitch\_x\), \(pitch\_y\)) in a real-world coordinate system, and their \textit{role}, \textit{team}, and \textit{jersey number}.
A detection is therefore represented as follows:

{\small{
\begin{equation} \label{hota}
d_{i}^{t} = \{ \underbrace{\textit{pitch\_x}, \textit{pitch\_y}}_{\text{localization}}, \underbrace{\textit{role}, \textit{team}, \textit{jersey\_number}}_{\text{identification}} \}.
\end{equation}
}}

While our main focus is football, the definition of the GSR task can extend to other team sports.

\section{SoccerNet-GSR Dataset} \label{sec:dataset}

Our dataset expands upon SoccerNet-Tracking~\cite{Cioppa2022SoccerNetTracking}, which consists of $200$ $30$-second clips split into train, validation, test, and a segregated challenge set.
In the original dataset, each frame includes bounding box annotations for the localization of players, referees, and balls tracked over time with extra role, team, and jersey number attributes.
Despite the comprehensive annotations, SoccerNet-Tracking lacks information like pitch localization, camera calibration\footnote{A camera calibration and pitch localization dataset was already introduced for the SoccerNet Camera Calibration challenge, but the corresponding annotations were provided on a separate set of data.}, and athlete positions on the pitch, critical for the Game State Reconstruction task.
In subsequent sections, we detail how we augmented the SoccerNet-Tracking annotations to create our proposed SoccerNet-GSR dataset.
The new annotations now include over $9.37$ million line points for pitch localization and camera calibration, as well as over $2.36$ million athlete positions on the pitch with their respective role, team, and jersey number.
Since the SoccerNet-GSR videos are uncut broadcast sequences captured by a single moving camera, only a portion of the football pitch is visible at any given time. 
As a result, the GSR task is limited to players within the camera's field of view.

\subsection{Athlete Localization on the Pitch}

Expressing the 2D image location of athletes in the real-world pitch coordinates requires pitch localization and camera calibration. 
Together, these information enable precise mapping of player positions from the image onto the pitch. 
Our new annotations described in this section therefore include: (1) pitch 2D positions, (2) camera parameters, and (3) positions on the pitch.


\mysection{Pitch localization.}
Following the same procedure as SoccerNet-v3~\cite{Cioppa2022Scaling}, we manually annotate every line on the football pitch by placing a series of points along its length to accurately define its shape, including curves such as the circles or the ones due to camera distortions. 
We categorize each line (\eg, side line left, side line top, \etc) and part of the goals, (left and right posts and the crossbar), totaling $26$ classes. 
Next, we continuously track all these annotations over time using key frame annotations and interpolations in-between when it is appropriate, mirroring the player tracking data as described in \cite{Cioppa2022SoccerNetTracking}, resulting in a densely marked pitch annotation throughout the entire video.
This annotation process is core for calibrating the camera through time.

\mysection{Camera calibration.}
Camera calibration is the process of determining the camera parameters for each frame, allowing to link the image-plane to the 3D world. It is required to compensate for the lack of a pre-calibrated camera.  Usually, this process requires correspondences between a known 3D object and its image. In the context of football, the pitch is a convenient object~\cite{IFAB2022Laws} to obtain correspondences from. In this work, we assume that the pitch has a conventional size of $105$ by $68$ meters. 
However, as the pitch is only partially visible in the images, the calibration of broadcast cameras is a challenging task. Hence, due to the lack of visible lines, some frames may not be calibrated correctly and are discarded in the evaluation. 
For the frames presenting a sufficient amount of pitch line annotations, 
the camera parameters are obtained from the best of several open-source techniques~\cite{Chen2018ATwoPoint,Magera2022SoccerNet} or an industrial tool~\cite{EVS2022Xeebra}. The complete process is described in the supplementary materials. 

\mysection{Position on the pitch.}
The point of calibrating the cameras is to derive positions in the real-world. Our 3D world reference axis system is centered on the pitch center mark, the X-axis points to the right goal, the Y-axis follows the middle line towards the camera and the Z-axis is perpendicular to the XY -- or the pitch -- plane, pointing towards earth's center.  
Once the camera parameters are known, 
the inverse of the camera projection function applied to a point gives a 3D ray that can be intersected with the pitch plane to derive the 3D position. We assume that the athlete's feet, and more specifically the center of the bottom part of their detection bounding boxes lies on the pitch. 
Unfortunately, this approximation limits the precision of the estimated locations in the case of jumps.
Hence, we remove the ball as it spends significant time in the air. 
A precise 3D localization of all elements would require tracking hardware, which is unavailable for open-science research at the moment.

\subsection{Athlete Identification} \label{sec:sports_persons_identification}
To identify athletes during a game, we leverage three distinct manual annotations provided in the SoccerNet-tracking dataset that have been previously overlooked in standard multi-object tracking: \textit{role}, \textit{team}, and \textit{jersey number}. 
The following paragraphs detail each annotation and the utilization of an additional \textit{track id} for cases where targets cannot be uniquely identified by their attributes.

\mysection{Role.}
In the SoccerNet-GSR dataset, athletes are categorized into four distinct roles during the game: 'player', 'goalkeeper', 'referee', or 'other'. 
The 'other' role encompasses individuals entering the pitch, such as coaches, medical staff, and any additional person not falling into the previous three categories.
For the first version of the SoccerNet-GSR benchmark, both referee responsibilities (i.e. main, bottom/top assistants) and ball detections are ignored.

\mysection{Team.}
Detections with the 'player' and 'goalkeeper' roles are annotated with a 'team' attribute, which can be assigned one of two values: 'left' or 'right'. 
Since the dataset consists of $30$-second sequences captured from a single camera, we determine the 'left' and 'right' teams based on their goal's position relative to the camera viewpoint.

\mysection{Jersey Number.}
Players and goalkeepers in the SoccerNet-GSR dataset are annotated with an additional 'jersey number' attribute.
However, unlike the role and team attributes, which are always available, a jersey number may not be visible at any point during the entire $30$-second video sequence.
In such cases, players with invisible shirt numbers are assigned a 'null' value for this attribute. 
If a player's jersey number is visible in at least one frame of the sequence, then the entire tracklet is annotated with that jersey number. 
Therefore, a jersey number assigned to a detection does not necessarily mean that it is visible in that particular frame.

\mysection{Track Id.}
We utilize the combination of role, team, and jersey number attributes to identify each athlete during a game. 
However, athletes cannot always be uniquely identified by their attributes. 
This occurs, for example, when two players from the same team do not have visible jersey numbers or when multiple individuals with the role of 'referee' or 'other' appear simultaneously.
Although these cases represent a small proportion of all annotated athletes, they prevent unique identification using attributes alone. 
To address this, we also include the standard 'track id' annotation from standard MOT. 
This requires methods for the SoccerNet-GSR task to output four values per detection: role, team, jersey number, and track id. 
The impact of non-uniquely identifiable targets is further discussed in \cref{sec:gs-hota}.

\section{GS-HOTA Evaluation Metric} \label{sec:gs-hota}
Game State Reconstruction (GSR) is a novel computer vision task closely related to multi-object tracking (MOT).
Yet, standard evaluation metrics for MOT, such as MOTA~\cite{Bernardin2008Evaluating} and HOTA~\cite{Luiten2020HOTA}, cannot be used to evaluate GSR for two main reasons.
First, these metrics do not account for additional attributes predicted on the tracked targets, such as team, role, and jersey numbers.
Second, these metrics rely on an IoU score to match predicted and ground truth bounding boxes in the image space, while GSR operates on 2D points within the pitch coordinate system.

To address these issues, we introduce \emph{GS-HOTA}, a novel evaluation metric to measure the ability of a GSR method to correctly track and identify all athletes on the sports pitch.
GS-HOTA is derived from the HOTA~\cite{Luiten2020HOTA} metric, 
which is formulated as follows:
{\small{
\begin{equation} \label{eq:hota}
\text{HOTA} = \int\limits_{\substack{0 < \alpha \leq 1}} \sqrt{\text{DetA}_\alpha \cdot \text{AssA}_\alpha}
\end{equation}
}}




\textit{DetA}/\textit{AssA} are the detection/association accuracy respectively, and $\alpha$ is a similarity threshold.
To compute these two underlying accuracy metrics, ground truth and predicted detections must first be matched according to a similarity score. 
Pairs with a similarity score below the $\alpha$ threshold are not matched.
For predictions (P) and ground truth (G) represented as bounding boxes in the image space, the Intersection-over-Union (IoU) is employed as the similarity score for the HOTA metric.
%
%
The key distinction setting GS-HOTA apart from HOTA is the use of a new similarity score, that accounts for the specificities of the GSR task, i.e. the additional target attributes (jersey number, role, team) and the detections provided as 2D points instead of bounding boxes.
This new similarity score, denoted $Sim_{GS-HOTA}(P, G)$, is formulated as follows:

\begin{equation} \label{eq:sim_gs_hota}
Sim_{\text{GS-HOTA}}(P, G) = \text{LocSim}(P, G) \times \text{IdSim}(P, G), \
\end{equation}




\begin{equation} \label{eq:locsim}
\text{with LocSim}(P, G) = e^{\ln(0.05)\frac{\|P - G\|_2^2}{\tau^2}}\ ,
\end{equation}

\begin{equation} \label{eq:idsim}
\text{and IdSim}(P, G) = 
\begin{cases} 
1 & \text{if all attributes match,} \\
0 & \text{otherwise.}
\end{cases}
\end{equation}

$Sim_{\text{GS-HOTA}}$, is therefore a combination of two similarity metrics.
The first metric, the localization similarity $\text{LocSim}(P, G)$, computes the Euclidean distance $\|P - G\|_2$ between prediction P and ground truth G in the pitch coordinate system. 
This distance is subsequently processed using a Gaussian kernel with a special \textit{distance tolerance parameter} $\tau$, resulting in a final score falling within the $[0, 1]$ range.
The second metric, the identification similarity $\text{IdSim}(P, G)$, is set to one only if all attributes match, i.e. role, team, and jersey numbers.
Attributes not provided in G are ignored, e.g. jersey numbers for referees\footnote{GSR methods must ignore the team and jersey number for non-player roles, as well as the jersey number when it is not visible in the video.}.
%
Finally, once P and G are matched, DetA and AssA are computed and integrated into a final GS-HOTA score, following the original formulation of the HOTA metric in \cref{eq:hota}.



\subsection{GS-HOTA Distance Tolerance Parameter}

\begin{figure}[t!]
\centering
\includegraphics[width=0.99\linewidth]{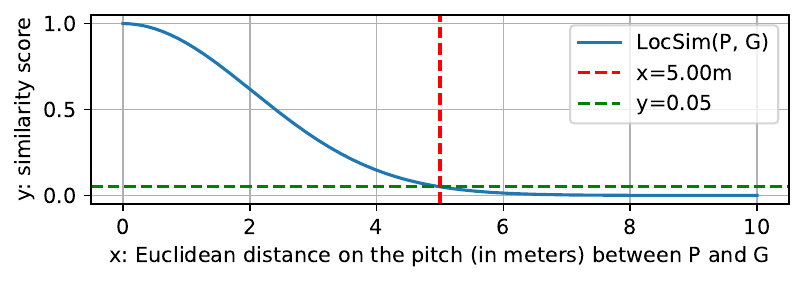}
  \caption{
    The localization similarity function for $\tau = 5$ meters.
  }
\label{fig:gaussian_sim_plot}
\end{figure}

Our GS-HOTA metric relies on a single $\tau$ parameter introduced in \cref{eq:locsim}. 
In practice, the continuous integral in \cref{hota} is computed over a discrete interval $\alpha \in [0.05, 0.95]$ with $0.05$ steps. 
This means that (P, G) pairs with a similarity below or equal to $0.05$ 
 are never matched.
Hence, our distance tolerance parameter $\tau$ defines the maximum distance in meters for a prediction P and a ground truth G to be matched, as illustrated in \cref{fig:gaussian_sim_plot}.
Furthermore, since all similarity thresholds in the range $[0.05, 0.95]$ are considered, a distance smaller than $\tau$ meters between P and G still results in a higher GS-HOTA.
This way, methods are still incentivized to produce athlete localization closer to the ground truth.
In this work, we define $\tau$ as $5$ meters, considering it a reasonable distance tolerance given the average dimensions of a soccer pitch ($68\times105$ meters) and the substantial distance between the camera and the athletes.

\subsection{Motivation and Discussion} \label{gsr-hota-motivations}
As introduced in \cref{sec:sports_persons_identification}, we consider the combination of attributes (role, team, jersey) as a way to identify athletes.
If each person was uniquely identifiable by the combination of its attributes, association would become trivial, and as a consequence, a simple average of the Detection Accuracy across all identities would suffice as a robust performance metric.
However, as explained in \cref{sec:sports_persons_identification}, not all identities in our SoccerNet GSR dataset can be uniquely identified by their attributes.
Therefore, the Association Accuracy must also be taken into account to account for identity switches among athletes sharing the same attributes (e.g. players from the same team with no visible jersey number).
%
%

Finally, a key difference that sets GSR apart from MOT — and by extension, GS-HOTA from HOTA — is the necessity to identify athletes by their attributes. 
This requirement is specified by \cref{eq:idsim}, according to which failing to correctly predict at least one attribute turns the corresponding detection into a False Positive.
Requiring the correct prediction of all attributes simultaneously is a strict constraint, which we justify based on the severe impact that incorrectly assigning localization data to a nonexistent or incorrect identity can have on downstream applications. 

\begin{figure*}[t!]
\centering
\includegraphics[width=0.99\linewidth]{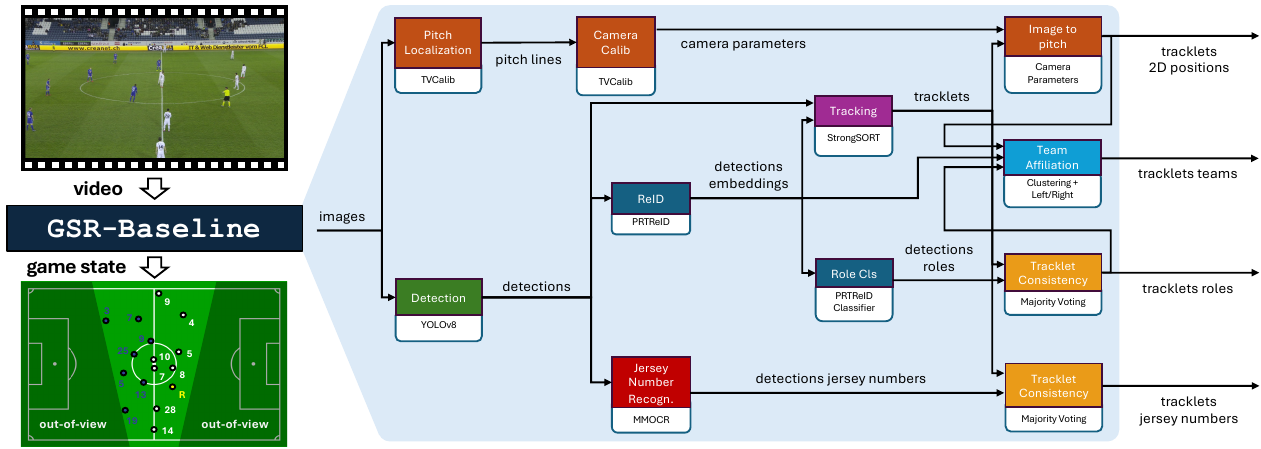}
  \caption{
  \textbf{Architecture overview of our proposed baseline.} 
  GSR-Baseline takes a video as input and outputs the complete game state. 
  Two modules are first applied on the input images: an object detector and a pitch localization model.  Then, PRTreID \cite{Mansourian2023Multitask} produces a ReID embedding for each detection, that is identity, team, and role aware. These embeddings are then forwarded to subsequent modules to perform role classification, team affiliation, and multi-object tracking. Finally, the pitch localization output is used for camera calibration, which enables the tracked bounding boxes to be transformed into 2D positions on the pitch coordinate system.
  }
\label{fig:architecture}
\end{figure*}

\section{GSR Baseline} \label{sec:baseline}
In this section, we introduce the GSR-Baseline, a pipeline designed to reconstruct the game state of any broadcast football video.
%
Our baseline splits the Game State Reconstruction task into several sub-tasks, selecting popular and open-source state-of-the-art methods for each sub-task.
To facilitate the development of such a complex video processing pipeline, we leverage TrackLab~\cite{Joos2024Tracklab}, a research-oriented PyTorch-based framework for multi-object tracking. 
The overall architecture of the GSR-Baseline is depicted in \cref{fig:architecture}, and a detailed description of each of the pipeline modules is provided hereafter.


\subsection{Athlete Detection and Tracking} \label{sec:baseline-tracking}
We employ a pre-trained \textbf{YOLOv8}~\cite{Jochter2023Yolov8} model as our athlete detector, without fine-tuning it on the SoccerNet dataset, since it already provides decent performance on football videos.
We filter the model's output to retain only the "person" class detections. 
To leverage existing strong multi-object trackers, our GSR-Baseline performs tracking in the image space based on bounding boxes.
As illustrated in \cref{fig:architecture}, these bounding boxes are converted into 2D pitch positions 
later within the pipeline.
Next, we employ \textbf{StrongSORT}~\cite{Du2023StrongSORT} as our multi-object tracker, for its SOTA performance and its ability to leverage both spatio-temporal and appearance cues, the latter being provided by the re-identification model PRTreID~\cite{Mansourian2023Multitask} described in \cref{sec:baseline-reid}.

\subsection{Pitch Localization and Camera Calibration} \label{sec:baseline-calib}

Camera calibration is performed using \textbf{TVCalib}~\cite{Theiner2023TVCalib}, which is composed of two modules. The first module performs pitch localization through semantic segmentation. The second estimates the camera calibration parameters by iteratively minimizing the pitch segments reprojection errors. 
Once the camera has been calibrated, its corresponding homography is used to transform image bounding boxes into 2D positions on the pitch. For this purpose, we assume that the bottom of the bounding box lies on the ground field.

\subsection{Athlete Identification} \label{sec:reid}
Athlete identification is performed by two key models: \textbf{PRTreID}~\cite{Mansourian2023Multitask} to produce team and role-aware ReID embeddings, and \textbf{MMOCR} for jersey number recognition. 
The output of these two models is further processed for tracklet consistency, team affiliation, and role classification, to produce the final game state identification data.

\mysection{Re-Identification.} \label{sec:baseline-reid}
The sportsperson representation model \textbf{PRTreID}~\cite{Mansourian2023Multitask} is designed to jointly solve person re-identification, role classification, and team affiliation with a single backbone.
Therefore, it produces an embedding that is team, role, and identity discriminative, thanks to a multi-task learning setup with three learning objectives.
PRTreID builds upon the SOTA part-based ReID method BPBreID~\cite{Somers2023Body}.
During the PRTreID training procedure, re-identification and team affiliation are formulated as deep metric learning tasks, where persons with the same identity/team are pulled close to each other in the embedding space with a triplet loss.
Role prediction is framed as a classification task with four target classes, employing a focal loss to address class imbalance.
At inference in the GSR-Baseline pipeline, PRTreID produces an embedding for each input detection, that is forwarded to subsequent modules to perform tracking, team clustering with left/right labeling, and role classification.





\mysection{Role Classification.} \label{sec:baseline-role}
The embeddings described above are processed by the \textbf{PRTreID} classification layer to output the target's role: \textit{player}, \textit{goalkeeper}, \textit{referee}, or \textit{other}.

\mysection{Jersey Number Recognition.}\label{sec:baseline-jn}
Jersey numbers recognition is performed in two separate steps with the open-source optical character recognition library \textbf{MMOCR}~\cite{Kuang2021MMOCR}. 
First, the YOLOv8 detections are fed to the \textbf{DBNet}~\cite{Liao2020RealTime} text detection model.
Subsequently, the detected texts are forwarded to the \textbf{SAR}~\cite{Li2019Show} text recognition model.
Finally, the highest-scored detected text containing a number is considered as the player's jersey number.



\mysection{Tracklet Consistency.} \label{sec:baseline-tracklet_consistency}
As described, jersey numbers and roles are predicted independently for each detection, potentially leading to inconsistencies within tracklets.
We adopt a \textbf{majority voting} approach within each tracklet to select the most common role and jersey number, ensuring uniformity. 

\mysection{Team Affiliation.} \label{sec:baseline-team}
Team affiliation is performed in three steps for tracklets having the "player" role assigned. 
First, the PRTReID embeddings of all detections within each tracklet are averaged to create a single tracklet-level representation of the player. 
Next, these tracklet-level embeddings are separated by a \textbf{K-means clustering} algorithm into two clusters representing two teams.
Finally, the average 2D positions of each team on the pitch are compared to determine which team is positioned more to the left or right. 

\section{Experiments}

\subsection{Implementation details}
%
To provide a baseline that is generic, we employ mostly pretrained networks that were not finetuned on SoccerNet. 
The only exceptions are PRTReid~\cite{Mansourian2023Multitask} and TVCalib~\cite{Theiner2023TVCalib}.
We use the standard weights provided by TVCalib's authors in our baseline.
Finally, PRTReid~\cite{Mansourian2023Multitask} is trained on the SoccerNet-GSR train set using parameters from the original paper.
For more implementation details, we invite readers to visit our project's GitHub repository and Tracklab\footnote{\url{https://github.com/TrackingLaboratory/tracklab}}.

\subsection{Evaluation}
To evaluate the performance of our proposed method on the Game State Reconstruction task, we employ the GS-HOTA metric introduced in \cref{sec:gs-hota}.
In the supplementary materials, we evaluate the performance of our GSR-Baseline in the image plane on the standard Multi-Object Tracking (MOT) task. 
Unless specified otherwise, all experiments are performed on the SoccerNet-GSR test set.

\subsection{Results and Analysis} \label{sec:results}

\begin{figure}[t!]
\centering
\includegraphics[width=0.99\linewidth]{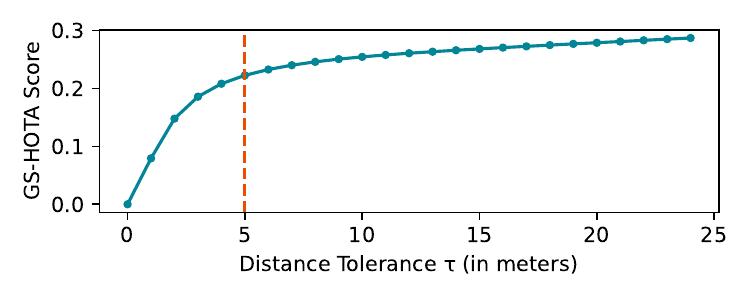}
  \caption{
  \textbf{Distance Tolerance Parameter $\tau$}: its influence on the GS-HOTA score.
  We pick $\tau$=5, illustrated by the orange line.
  }
\label{fig:gshota_vs_tolerance}
\end{figure}

\mysection{Main Results and GS-HOTA Analysis.}
We report the performances of our GSR-Baseline in \cref{table:gs-hota}, which achieves $22.26\%$ in GS-HOTA on the test set.
All experiments in this table correspond to slight variations of the $Sim_{\text{GS-HOTA}}(P, G)$ introduced in \cref{eq:sim_gs_hota}.
First, when ``Pitch'' is disabled, the \textit{LocSim} function in \cref{eq:locsim} is replaced with the bounding boxes IoU in the image space: pitch localization and camera calibration have therefore no impact.
Second, we ablate each attribute of the identification component in \cref{eq:idsim} (\textit{IdSim} is set to 1 when all attributes are disabled). 
The first experiment in \cref{table:gs-hota} falls back to the standard HOTA, \ie with the IOU in image space as a similarity function.
The remaining experiments illustrate how enabling attributes in \cref{eq:idsim} induces successive drops in performance, since it introduces additional predictions in the evaluation and therefore potential errors.
\cref{table:gs-hota} also highlights the key challenges of this task, showing that our GSR-Baseline struggles mostly with jersey number recognition, followed by pitch localization, team affiliation, and finally role classification.
Finally, the influence of the GS-HOTA distance tolerance parameter $\tau$ introduced in \cref{sec:gs-hota} is illustrated in \cref{fig:gshota_vs_tolerance}.
According to this plot, picking $\tau=5$ meters is a reasonable choice since performance quickly drops with a stricter tolerance.
\begin{table}[]
    \centering
    \caption{
    \textbf{Main Results and GS-HOTA Analysis.}
    Attributes (Role, Team, Jersey) are ignored in the GS-HOTA computation when disabled.
    IoU in image space is used when Pitch is disabled.
    }
    \label{table:gs-hota}
    \resizebox{0.925\columnwidth}{!}{
    \begin{tabular}{lccccr}
    \toprule
     \multirow{2}{*}{Split} & \multicolumn{4}{c}{GS-HOTA components} & \multirow{2}{*}{GS-HOTA $\uparrow$} \\ \cmidrule{2-5}
    & Pitch     & Role     & Team    & Jersey     &                          \\ \midrule 
    \multirow{7}{*}{Test} &   \xmark                &  \xmark        &   \xmark      &  \xmark                & 57.64                  \\
    & \cmark         &  \xmark        & \xmark  &  \xmark           &  42.65                 \\ 
    & \cmark         & \cmark        &  \xmark       &   \xmark        &  40.76                 \\ 
    & \cmark         &  \xmark     & \cmark       &  \xmark         &  37.03                 \\ 
    & \cmark         &  \xmark     &   \xmark         & \cmark       & 25.65                  \\ 
    & \xmark             & \cmark        & \cmark       & \cmark       & 29.50                  \\ 
    & \cmark         & \cmark        & \cmark       & \cmark       & \textbf{22.26} \\
    \midrule
    Valid    & \cmark         & \cmark        & \cmark       & \cmark       & \textbf{18.05} \\
    \midrule
    Challenge    & \cmark         & \cmark        & \cmark       & \cmark       & \textbf{23.36} \\
    \bottomrule
    \end{tabular}
    }
\end{table}

\mysection{Ablation Study of GSR-Baseline Modules.} \label{sec:baseline-analysis}
\Cref{table:baseline-analysis} illustrates the impact of each module on the overall performance. 
This study employs the ground truth as an oracle for all modules except the module of interest and its downstream modules in the pipeline. 
For instance, when examining the ReID module, the tracking, role classification, and team clustering modules are also activated. 
Dependencies between modules are depicted as a flowchart in \cref{fig:architecture}.
The first experiment (Exp. 1) shows that the heuristic chosen for team 'left'/'right' affiliation is highly effective, especially considering the significant impact that swapping two teams can have on GS-HOTA.
Similarly, Exp. 2 demonstrates the solid performance of all modules depending on the ReID embeddings (\ie tracking, role cls, and team aff.).
Furthermore, Exp. 3 and 4 show the severe performance impact of enabling calibration and pitch localization, suggesting ample opportunities for improvements with these two modules.
%
Similarly, Exp. 5 with the jersey number recognition module exposes it as another key weakness of the pipeline. 
%
Finally, performance in Exp. 6 is close to the complete baseline, since the object detector is the starting point for most of the pipeline, and ground truth data is therefore employed here only for pitch localization and camera calibration. 
%
%

Our ablation study shows that while localization and identification are challenging alone, their intricate combination in GSR proves even more challenging.

\begin{figure*}[ht!]
     \centering
     \begin{subfigure}[b]{0.35\textwidth}
         \centering
         \includegraphics[width=\textwidth]{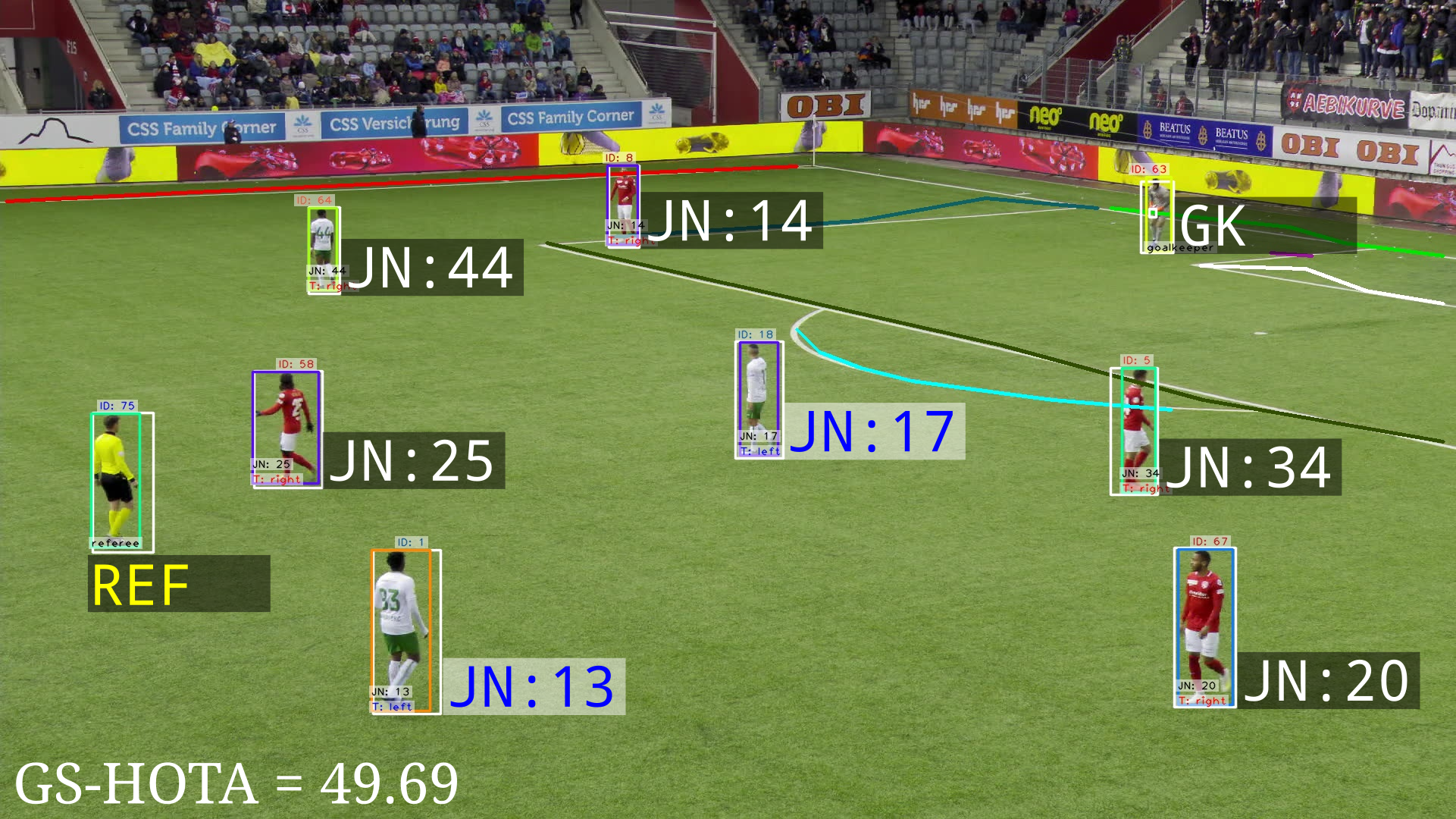}
     \end{subfigure}
     \begin{subfigure}[b]{0.28\textwidth}
         \centering
         \includegraphics[width=\textwidth]{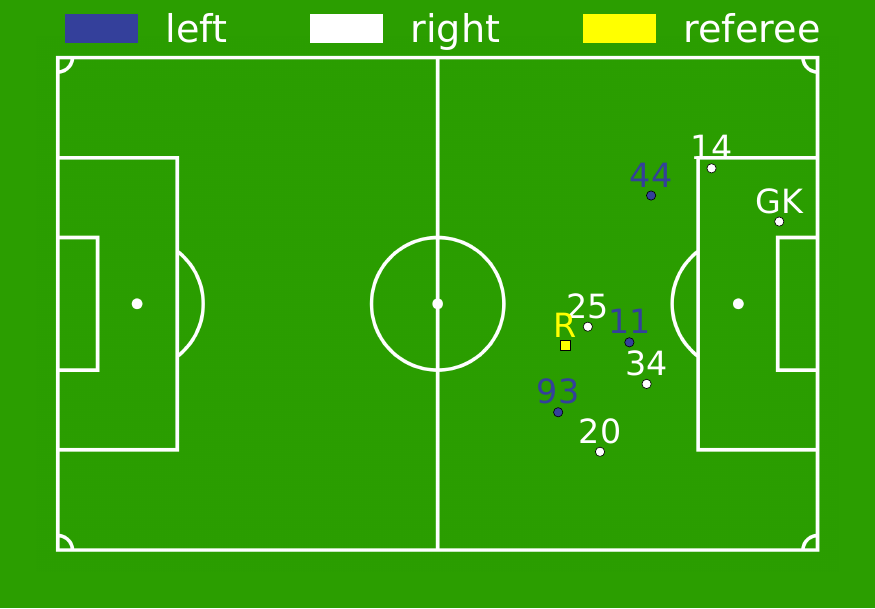}
     \end{subfigure}
     \begin{subfigure}[b]{0.28\textwidth}
         \centering
         \includegraphics[width=\textwidth]{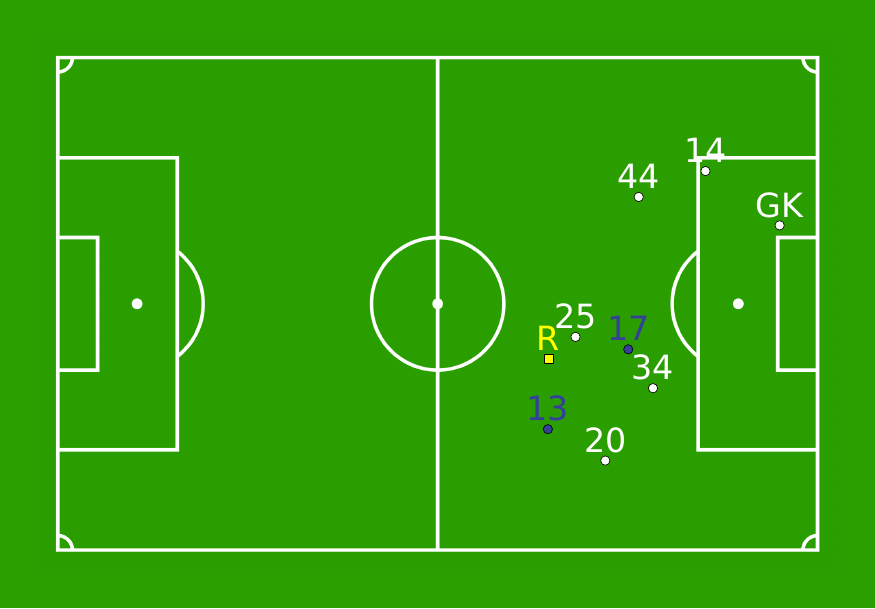}
     \end{subfigure}
     \begin{subfigure}[b]{0.35\textwidth}
         \centering
         \includegraphics[width=\textwidth]{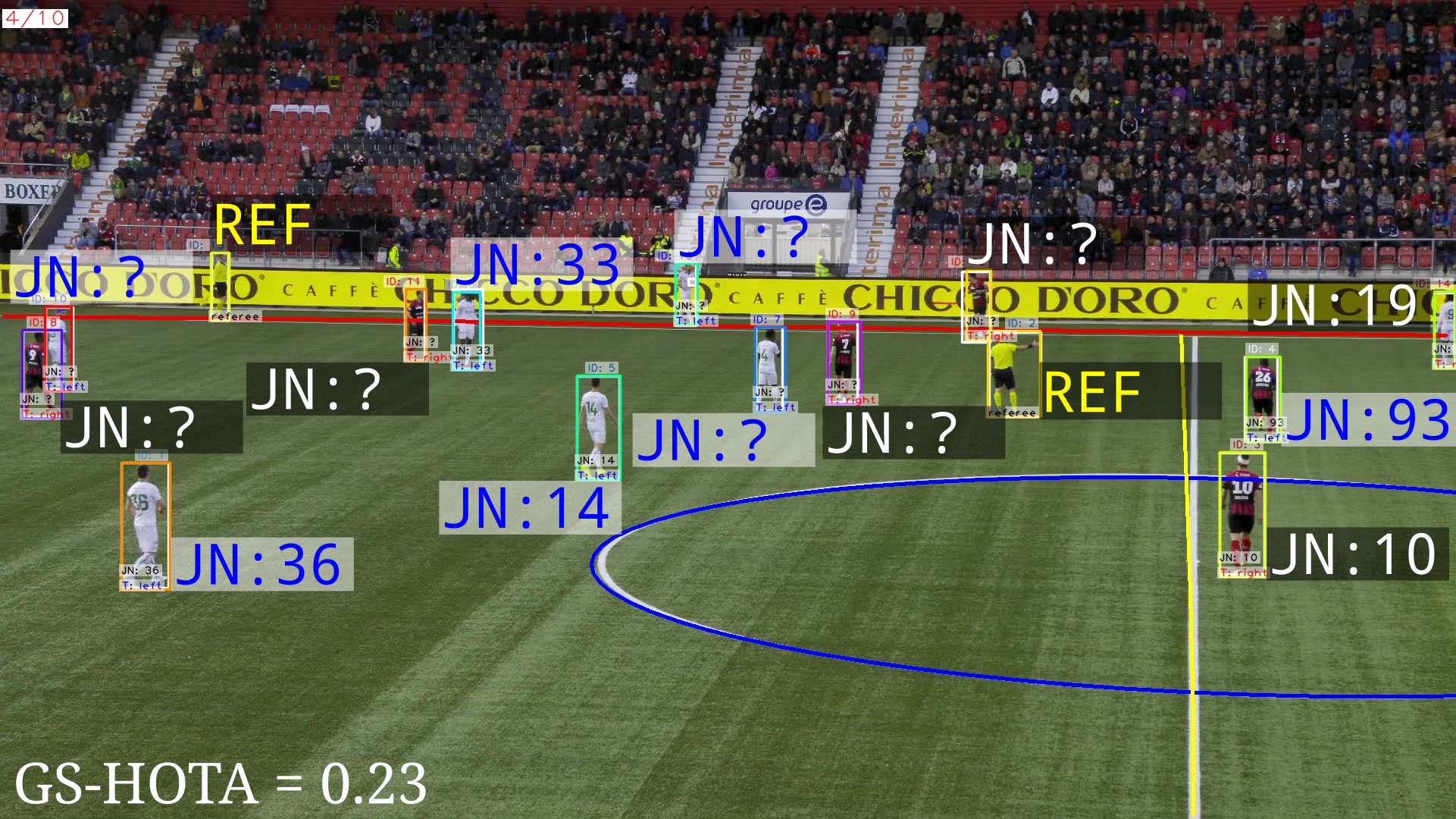}
         \caption{Predictions on input image}
     \end{subfigure}
     \begin{subfigure}[b]{0.28\textwidth}
         \centering
         \includegraphics[width=\textwidth]{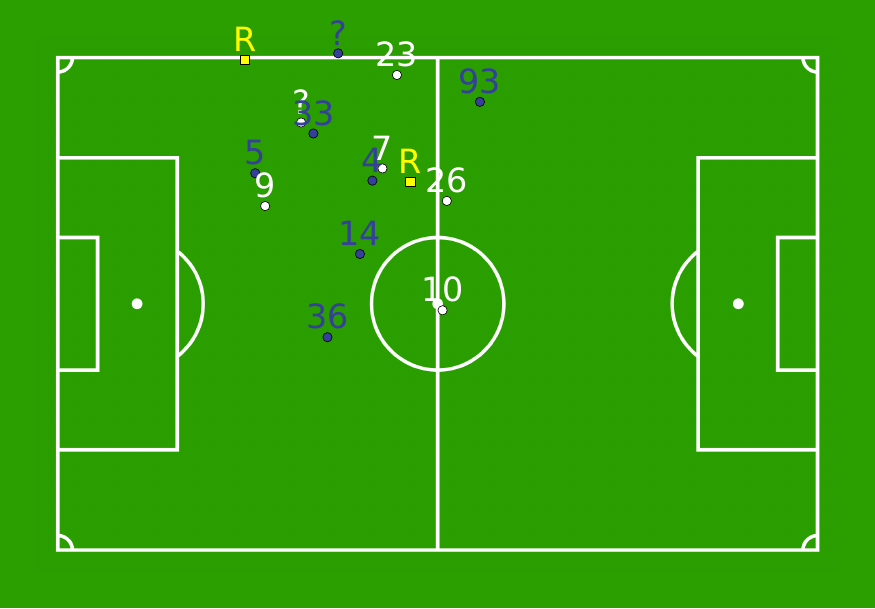}
         \caption{Ground Truth minimap}
     \end{subfigure}
     \begin{subfigure}[b]{0.28\textwidth}
         \centering
         \includegraphics[width=\textwidth]{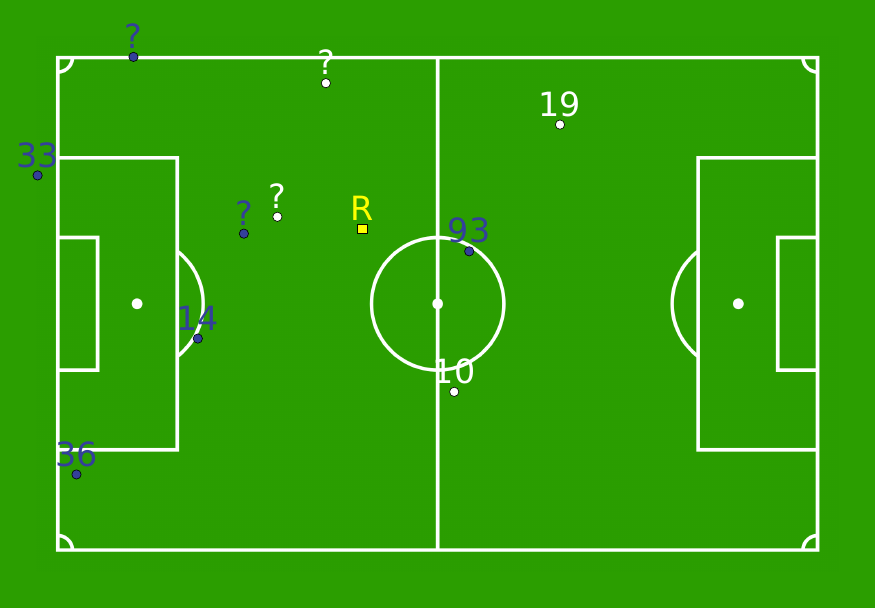}
         \caption{Predicted minimap}
     \end{subfigure}
        \caption{\textbf{Qualitative results.} 
        Output predictions of two frames from videos with different GS-HOTA values. (Top) High GS-HOTA (49.69\%), with robust pitch localization and accurate athlete identification.
        (Bottom) Calibration failure (\eg due to insufficient pitch elements) leads to completely erroneous athlete localization and poor GS-HOTA (0.23\%).
        }
        \label{fig:qualitative}
\end{figure*}


\newcommand{\cm}{\checkmark}
\begin{table}[]
    \centering
\caption{
\textbf{GSR-Baseline Ablation Study.} 
We report the GS-HOTA for each GSR-Baseline module and its corresponding downstream modules by replacing other modules by a ground truth oracle.
We also report their speed in FPS and their input batch sizes.
}
\label{table:baseline-analysis}

\resizebox{\columnwidth}{!}{
\begin{tabular}{rlrrr}\toprule
& Module             & GS-HOTA $\uparrow$ & Batch S. & FPS \\ \midrule
{\footnotesize (1)} & Team Side       & 92.00 & Video & 1.5K  \\
{\footnotesize (2)} & ReID {\footnotesize (PRTReID)}        & 87.42 & 16  & 14.5 \\
{\footnotesize (3)} & Calibration {\footnotesize (TVCalib)}    & 51.39 & 512 & 7.6  \\
{\footnotesize (4)} & Pitch {\footnotesize (TVCalib)}         & 49.99 & 16  & 2.9   \\
{\footnotesize (5)} & Jersey N° {\footnotesize (MMOCR)}   & 56.75 & 32  & 3.8   \\
{\footnotesize (6)} & BBox Det. {\footnotesize (YOLOv8)}  & 35.28 & 32  & 16.5  \\\midrule
& Full Baseline   & 22.26 & N/A & 1.1 \\\bottomrule
\end{tabular}}
\end{table}

\mysection{Inference Time.}
%
Since the GSR-Baseline is an offline pipeline, each module processes its input in batches, where a single batch can span multiple images.
The batch size and average frame rate of each module are reported in \cref{table:baseline-analysis}.
All inference speed tests are performed with an NVIDIA A100 32GB GPU.
As illustrated, pitch localization, camera calibration, and jersey-number recognition emerge as the most time-consuming modules.
It takes on average $11$ minutes to process one $30$s sequence from our dataset.

\mysection{Qualitative Results.}
\cref{fig:qualitative} illustrates two game state minimaps predicted by our GSR-Baseline and their respective ground truths.
Our GSR-Baseline achieves a high GS-HOTA score of $49.69\%$ on the video illustrated in the first row, accurately predicting most athletes' pitch positions and attributes.
The bottom example, from a video with a GS-HOTA of $0.23\%$, exemplifies common failure cases, where even minor calibration inaccuracies can cause major pitch registration errors. 
In this frame, poor calibration is caused by the small number of visible salient points on the pitch.

\section{Conclusion}
Our work introduces the first Game State Reconstruction (GSR) benchmark for athlete identification and tracking on a minimap, comprising a new dataset, evaluation metric, and open-source baseline. 
Unlike previous efforts in sports video understanding that focused on specific subtasks, our approach stands out by benchmarking a complete pipeline, whose high-level game semantics outputs are directly relevant to a broad spectrum of downstream applications. 
Moreover, experiments with our proposed baseline reveal the inherent complexity of the GSR task and the significant interdependencies among its various subtasks. 
We hope that our introduced benchmark will pave the way for a new line of exciting research on specialized GSR methods.
We anticipate future efforts to focus on (1) enhancing specific modules to increase performance, (2) implementing real-time pipelines, or even (3) developing end-to-end differentiable methods for tackling the task in one step.

\paragraph*{Acknowledgments.}


This work was supported by SportRadar, the Service Public de Wallonie (SPW) Recherche, under the ReconnAIssance project and Grant $\text{N}^{\text{o}}$8573, the F.R.S-FNRS, FRIA/FNRS, the King Abdullah University of Science and Technology (KAUST) Office of Sponsored Research through the Visual Computing Center (VCC) funding and the SDAIA-KAUST Center of Excellence in Data Science and Artificial Intelligence (SDAIA-KAUST AI). 
Computational resources have been provided by the supercomputing facilities of the Université catholique de Louvain (CISM/UCLouvain).

{
    \small
    \bibliographystyle{ieeenat_fullname}
    \bibliography{bib/abbreviation-short, 
    bib/activity, 
    bib/camera-calibration-sports,
    bib/dataset,
    bib/deepsport,
    bib/detection,
    bib/explainability, 
    bib/labo, 
    bib/learning, 
    bib/library,
    bib/llm, 
    bib/multiview, 
    bib/online, 
    bib/referee, 
    bib/re-identification,
    bib/soccer, 
    bib/soccernet-challenge, 
    bib/sports,
    bib/tracking}

\clearpage
\setcounter{page}{1}
\maketitlesupplementary
\setcounter{section}{0}
\renewcommand\thesection{\Alph{section}}
\setcounter{figure}{0}
\let\oldthefigure\thefigure
\renewcommand{\thefigure}{S\oldthefigure}

\setcounter{table}{0}
\let\oldthetable\thetable
\renewcommand{\thetable}{S\oldthetable}

\section{Camera calibration}
The estimated camera parameters follow the pinhole camera model augmented with one radial distortion coefficient that may be needed in wide camera shots.
The camera parameters are estimated in four steps. First, as a global pre-processing step, the intersections of the pitch markings are computed to obtain both line-to-line and point-to-point correspondences between the image and the soccer pitch model.
Then, depending on the visible parts of the soccer pitch in the image, different strategies are used to retrieve camera parameters. When there is a sufficient amount of pitch markings in an image, a homography mapping the image plane to the soccer pitch plane is estimated, then converted into pinhole camera parameters. Moreover, an optimization is conducted to determine one radial distortion coefficient given the curvature of the annotated polylines. 
For the frames that do not display enough pitch markings, the knowledge that each sequence is shot by a single camera is leveraged.  As broadcast cameras are both zooming and rotating, only the camera position can be considered fixed.  It is estimated as the median 3D position of the camera parameters estimated in the previous step. A similar version of the Two-Point PTZ algorithm is used to compute the focal length, pan and tilt parameters. 
Finally, as there are still some frames that can not be calibrated with sufficient accuracy, an industrial tool is used to compute the camera parameters of the missing frames. 

Some examples of the pitch annotations used for the camera calibration can be found in \cref{fig:pitch_annotations}.

\begin{figure}[ht]
    \centering
    \includegraphics[width=0.99\linewidth]{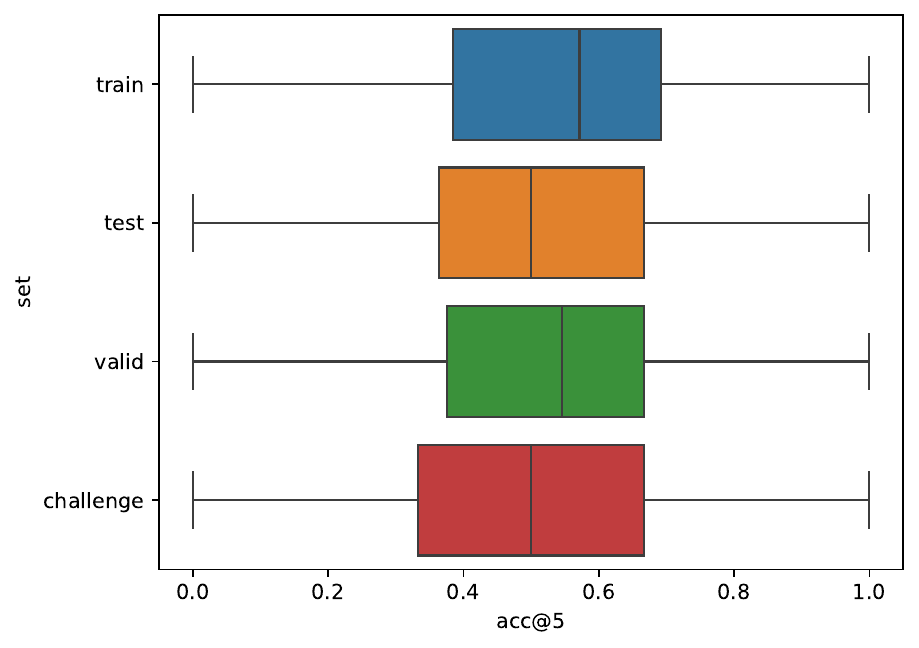}
    \caption{Distribution of the \texttt{acc@5} metric for the different sets}
    \label{fig:enter-label}
\end{figure}

\begin{figure*}[ht!]
     \centering
     \begin{subfigure}[b]{0.33\textwidth}
         \centering
         \includegraphics[width=\textwidth]{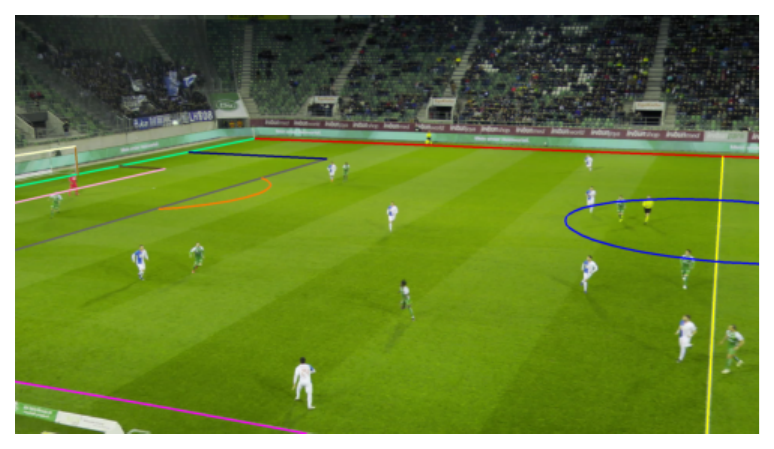}
     \end{subfigure}
     \begin{subfigure}[b]{0.33\textwidth}
         \centering
         \includegraphics[width=\textwidth]{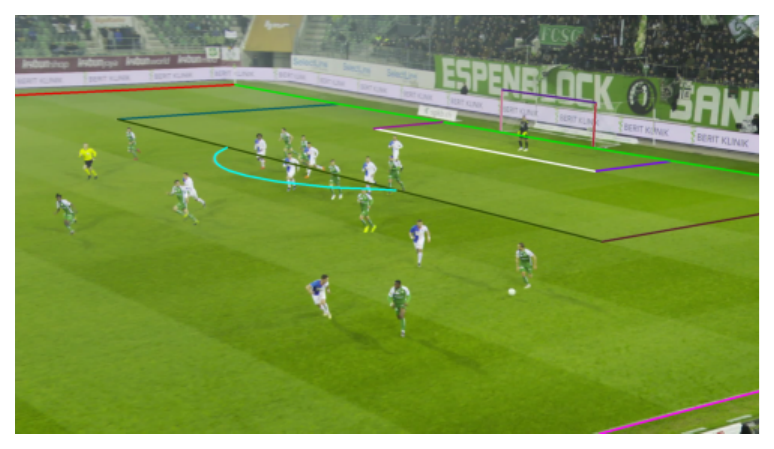}
     \end{subfigure}
     \begin{subfigure}[b]{0.33\textwidth}
         \centering
         \includegraphics[width=\textwidth]{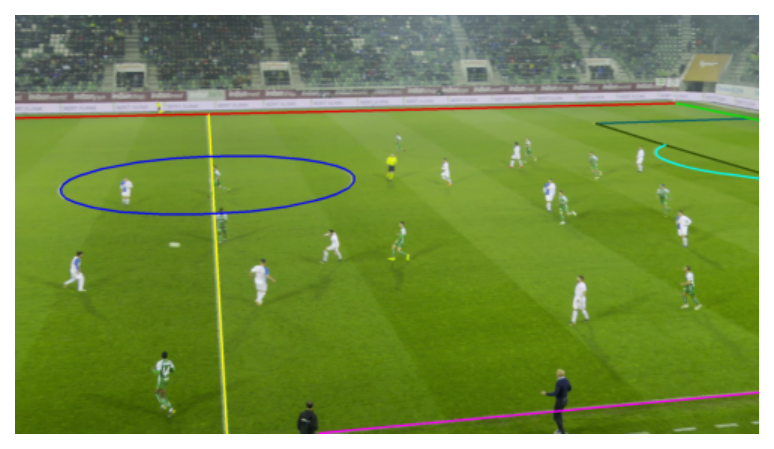}
     \end{subfigure}
     \begin{subfigure}[b]{0.33\textwidth}
         \centering
         \includegraphics[width=\textwidth]{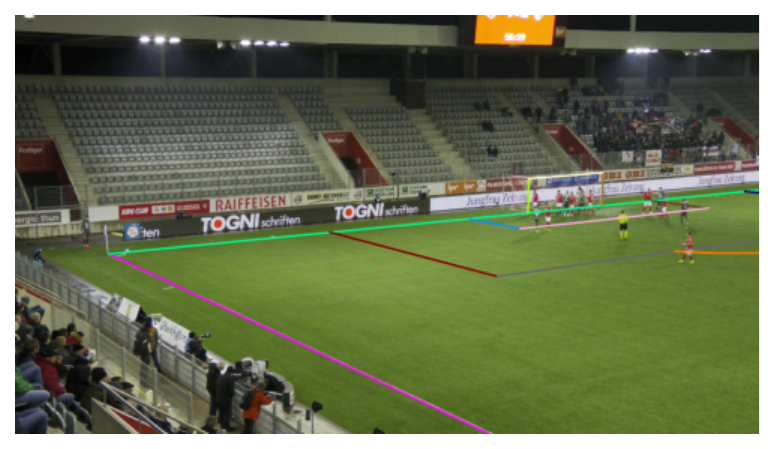}
     \end{subfigure}
     \begin{subfigure}[b]{0.33\textwidth}
         \centering
         \includegraphics[width=\textwidth]{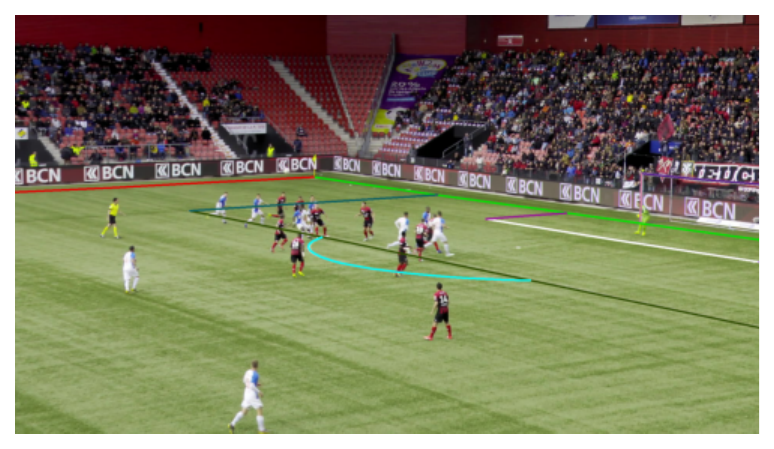}
     \end{subfigure}
     \begin{subfigure}[b]{0.33\textwidth}
         \centering
         \includegraphics[width=\textwidth]{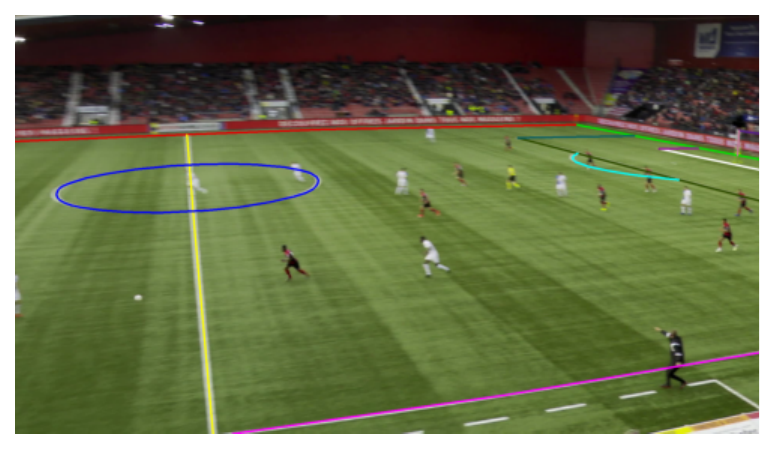}
     \end{subfigure}
        \caption{\textbf{Pitch annotations.} 
        Examples of pitch annotations.
        }
        \label{fig:pitch_annotations}
\end{figure*}

\section{GS-HOTA additional discussion}

The HOTA authors introduced a ``Classification-Aware HOTA" that shares similarities with our proposed GS-HOTA.
However, the Classification-Aware HOTA is not suitable for evaluating game state reconstruction for several reasons: first, it imposes a less rigid constraint on class predictions than \cref{eq:idsim}, second, it is tailored for a single classification objective, and third, it necessitates to output one classification score for each potential class, all summing up to one, an unsuitable requirement for team affiliation and jersey number recognition. 

\section{Comparison with Standard MOT Methods.}
We present the performance of the `image tracking only' component of our baseline, which includes the detector, ReID, and tracking modules, to compare with existing SOTA Multi-Object Tracking (MOT) methods.
To this end, we employ two well-established metrics: MOTA~\cite{Bernardin2008Evaluating} and HOTA~\cite{Luiten2020HOTA}.
Results in \cref{table:mot} reveal our superior performance over methods using non-fine-tuned object detectors.
Finally, a specialized soccer tracking method such as \cite{Maglo2023Individual} highlights the potential for improvement in image-based tracking.
This method relies on a strong object player detector fine-tuned on soccer data, and a heavy test-time fine-tuning of a ReID model to associate short tracklets into long tracks and achieve long-term tracking.


\begin{table}[t]
    \centering
    \caption{
    \textbf{GSR-Baseline on SoccerNet Tracking~\cite{Cioppa2022SoccerNetTracking}.}
    \textit{FT Det} indicate an object detector fine-tuned on SoccerNet.
    GSR-B uses an out-of-the-box YOLOv8, not fine-tuned on soccer data.
    }
    \label{table:mot}
    \resizebox{\linewidth}{!}{
    \begin{tabular}{lcrrrr} 
        \toprule
        Algorithm     & FT Det &   HOTA    &   DetA    &   AssA    &   MOTA      \\         \midrule 
        DeepSORT \cite{Wojke2017Simple} &      &   36.66  &   40.02  &   33.76  &   33.91  \\       
        FairMOT \cite{Zhang2021FairMOT} &       &   43.91  &   46.32  &   41.78  &   50.70   \\         
        ByteTrack \cite{Zhang2022Bytetrack} &      & 47.23    &	44.49  &	50.26  &   31.74     \\        
        GSR-B (ours) &    &  \bf 57.64  &      \bf67.42      &      \bf 49.42     &   \bf 80.79    \\      
        \midrule
        FairMOT-ft \cite{Zhang2021FairMOT} & \checkmark    &  57.88  & 66.56  &  50.49  &  83.56    \\
        SNT23-Winners \cite{Maglo2023Individual} & \checkmark    &  \bf 73.29  &   \bf 73.26  &  \bf  73.42  &  \bf  87.74    \\     
        \bottomrule
    \end{tabular}
    }
\end{table}

\section{Annotation sample}
An annotation sample in JSON format is illustrated in \cref{fig:annotations} for a single video.
\begin{figure*}[t!]
\centering
\begin{lstlisting}[]
{
    "info":{
      "version":"1.1",
      "game_id":"11",
      "id":"200",
      "num_tracklets":"20",
      "action_position":"956196",
      "action_class":"Shots on target",
      "visibility":"visible",
      "game_time_start":"2 - 15:41",
      "game_time_stop":"2 - 16:11",
      "clip_start":"941000",
      "clip_stop":"971000",
      "name":"SNGS-200",
      "im_dir":"img1",
      "frame_rate":25,
      "seq_length":750,
      "im_ext":".jpg"
    },
    "images": [
      {
         "is_labeled":true,
         "image_id":"3200000001",
         "file_name":"000001.jpg",
         "height":1080,
         "width":1920,
         "has_labeled_person":true,
         "has_labeled_pitch":true,
         "has_labeled_camera":true,
         "ignore_regions_y":[],
         "ignore_regions_x":[]
      },
      // Additional images annotations...
     ],
    "annotations":[
       {
          "id":"3200000001",
          "image_id":"3200000001",
          "track_id":1,
          "supercategory":"object",
          "category_id":1,
          "attributes":{
             "role":"player",
             "jersey":"14",
             "team":"left"
          },
          "bbox_image":{
             "x":1020,
             "y":508,
             "x_center":1043.0,
             "y_center":557.5,
             "w":46,
             "h":99
          },
          "bbox_pitch":{
             "x_bottom_left":-29.17307773076183,
             "y_bottom_left":-13.960906317008366,
             "x_bottom_right":-28.399824812615115,
             "y_bottom_right":-14.278786952621587,
             "x_bottom_middle":-28.786446826184775,
             "y_bottom_middle":-14.119801608871501
          }
       },
       // Additional athletes annotations...
       ...
\end{lstlisting}
\end{figure*}

\begin{figure*}[t!]
\centering
\begin{lstlisting}[firstnumber=66]
       ...
       {
        "id":"3200000019",
        "image_id":"3200000001",
        "supercategory":"pitch",
        "category_id":5,
        "lines": {
          "Side line top":[{"x":0.21, "y":0.34}, {"x":0.61, "y":0.39}, {"x":1.0, "y":0.43}],
          "Side line left":[{"x":0.0, "y":0.45}, {"x":0.10, "y":0.39}, {"x":0.21, "y":0.34}],
          "Small rect. left top":[{"x":0.07, "y":0.53}, {"x":0.01, "y":0.53}, {"x":0.01, "y":0.53}],
          "Small rect. left main":[{"x":0.0, "y":0.54}, {"x":0.01, "y":0.54}, {"x":0.01, "y":0.53}],
          "Big rect. left top":[{"x":0.04, "y":0.42}, {"x":0.23, "y":0.45}, {"x":0.41, "y":0.48}],
          "Big rect. left main":[{"x":0.0, "y":0.81}, {"x":0.20, "y":0.65}, {"x":0.41, "y":0.48}],
          "Circle left":[{"x":0.02, "y":0.79}, {"x":0.04, "y":0.79}, ...],
        }
      }
      // Additional pitch annotations...
    ]
}
\end{lstlisting}
\caption{Sample JSON annotation for one video of the SoccerNet-GSR dataset}
\label{fig:annotations}
\end{figure*}

\end{document}